\documentclass[letterpaper, 10pt, conference]{ieeeconf}      % Use this line for a4 paper
\usepackage{dirtytalk}
\usepackage{longtable}
\usepackage{cite}
\usepackage{lipsum,booktabs}
\usepackage[table,x11names]{xcolor}
\usepackage{amsmath,array,graphicx}
\usepackage{tabularx,booktabs}
\usepackage{stfloats}
\usepackage{float}
\usepackage{lipsum,multicol}
\usepackage{upquote}

\IEEEoverridecommandlockouts                              % This command is only needed if 
\title{\LARGE \bf
Age-Appropriate Robot Design: In-The-Wild Child-Robot Interaction Studies of Perseverance Styles and Robot's Unexpected Behavior
}

\author{Alicja Wróbel$^{1}$, Karolina Źróbek$^{2}$, Marie-Monique Schaper$^{3}$, Paulina Zguda$^{4}$ and Bipin Indurkhya$^{5}$ 
\thanks{This research was supported in part by a grant from the Priority Research Area DigiWorld PSP: U1U/P06/NO/02.19 under the Strategic Programme Excellence Initiative at the Jagiellonian University, and by the National Science Centre, Poland, under the OPUS call in the Weave programme under the project number K/NCN/000142.}% <-this % stops a space
\thanks{$^{1}$Alicja Wróbel is a Masters Degree Student at the Faculty of Philosophy, Jagiellonian University in Cracow, Poland.
        {\tt\small alka.wrobel@student.uj.edu.pl}}%
\thanks{$^{2}$Karolina Źróbek is a Bachelors Degree Student at the Faculty of Mathematics and Computer Science, Jagiellonian University in Cracow, Poland.
        {\tt\small karolina.zrobek@student.uj.edu.pl}}
\thanks{$^{3}$Marie-Monique Schaper is at the Center for Computational Thinking and Design at Aarhus University, Denmark.
{\tt\small mmschaper@cc.au.dk}}
\thanks{$^{4}$Paulina Zguda is a PhD Student at the Faculty of Philosophy, Jagiellonian University in Cracow, Poland.
{\tt\small paulina.zguda@doctoral.uj.edu.pl}}
\thanks{$^{5}$Bipin Indurkhya is at the Cognitive Science Department at Jagiellonian University in Cracow, Poland.
{\tt\small bipin.indurkhya@uj.edu.pl}}
}

\begin{document}

\maketitle
\thispagestyle{empty}
\pagestyle{empty}

%%%%%%%%%%%%%%%%%%%%%%%%%%%%%%%%%%%%%%%%%%%%%%%%%%%%%%%%%%%%%%%%%%%%%%%%%%%%%%%%
\begin{abstract}
As child-robot interactions become more and more common in daily life environment, it is important to examine how robot's errors influence children's behavior. We explored how a robot's unexpected behaviors affect child-robot interactions during two workshops on active reading: one in a modern art museum and one in a school. We observed the behavior and attitudes of 42 children from three age groups: 6-7 years, 8-10 years, and 10-12 years. Through our observations, we identified six different types of surprising robot behaviors: personality, movement malfunctions, inconsistent behavior, mispronunciation, delays, and freezing. Using a qualitative analysis, we examined how children responded to each type of behavior, and we observed similarities and differences between the age groups. Based on our findings, we propose guidelines for designing age-appropriate learning interactions with social robots.
\end{abstract}

%%%%%%%%%%%%%%%%%%%%%%%%%%%%%%%%%%%%%%%%%%%%%%%%%%%%%%%%%%%%%%%%%%%%%%%%%%%%%%%%
\section{Introduction}
As the use of social robots with children becomes increasingly common, research has focused on their potential as playmates, caregivers, educators, and therapists \cite{Tolksdorf2021} \cite{newton_humanoid_2019} \cite{DiPietro2019}. However, it is important to understand children's attitudes towards social robots, and several studies have examined this \cite{Zguda2021} \cite{Straten2020} \cite{bjorling2019effect}.

Unexpected situations in child-robot interactions can arise due to the robot's appearance, behavior, and system crashes \cite{serholt_trouble_2020}. Sometimes, these unexpected situations may positively influence children. For example, Lemaignan et al. \cite{Lemaignan2015} found that children were more interested in a misbehaving robot than in one that behaved predictably. Yadollahi et al. \cite{yadollahi_when_2018} suggest that a robot's mistakes can enhance children's participation in the interaction and promote learning by allowing the child to correct the robot's inaccuracies. However, unexpected behavior can also have negative effects, such as decreasing the child's trust in the robot \cite{geiskkovitch_what?_2019} and may even generate fear\cite{Rubegni2022}. Children have expressed fear related to robot's deceptive behaviors, potential lack of control, and unpredictability of its actions \cite{yadollahi_when_2018}. Rubegni et al. \cite{Rubegni2022} found that children also fear unpredictable emotional reactions from the robot, as well as breakdowns that could harm people or unintentional strength lacking control. Gamboa \cite{Gamboa2022} reports on critical incidents and implications of use when using social robots with young children at home. Although there have been many studies on children's perceptions of social robots over the past years, most of them focus on understanding children's emotions and behavior of one specific age group and less on comparing age-related differences \cite{yadollahi_when_2018}. In this regard, Wang et al. \cite{wang_informing_2022} stress the need for designing age-appropriate interactions between children and artificial intelligence, because children's needs vary as they develop cognitively. 

To address this need, we explored the design of age-appropriate learning experiences with social robots for children, taking into account the children's age differences. Specifically, We focused on the research question: {\it How to design age-appropriate learning experiences for children in natural settings with social robots, taking into account their positive and negative reactions to unexpected behaviors?} Our study involved 42 children aged 6-12 years old, and our findings highlight the influence of the robot's unexpected behavior on children's responses during child-robot interaction and their sense of safety. We propose design implications based on these findings. 

\section{Study design}
\subsection{In-the-wild methodology}
The study was conducted following the in-the-wild methodology, where a phenomenon is studied in its natural environment instead of in a laboratory. To design better real-life interactions between children and robots in the future, it is necessary to examine children's perceptions of robots in their natural environment while they are engaged in a familiar task. Studies conducted in laboratories lack major social and contextual factors influencing children's behavior while interacting with a robot \cite{crabtree2020research}.

\subsection{Setup of the study}
We conducted four workshop sessions in this study in Cracow, Poland. The first two sessions were held at the Museum of Contemporary Art in Cracow, and the third and the fourth sessions were held at an international school. The children that participated in the first session (10 children) were of 10-12 years, the ones in the second session (8 children) were of 6-7 years, the ones in the third session (12 children) of 11-12 years, and the ones in the fourth session (12 children) of 8-10 years. These age groups were selected on the basis of those interested in joining the event, as well as the availability of the facilities and staff concerned to carry out the interaction. The sessions one, three, and four were conducted in English, and the session two was conducted in Polish. Any child whose first language was not the dominant language in any session was given ongoing support and translation by the researchers. All the children were familiar with the venues of the study: they either attended the school where the workshop was held, or had participated in extracurricular activities at the museum before.

The procedure involved reading a book, by NAO and the facilitator, to the children with intermittent crafting, motor, or verbal activities to emotionally engage with the text and enhance understanding. During group interactions with the robot, children could converse with it and ask questions. During this part, the robot was controlled using the Wizard-of-Oz methodology, where a researcher answers the children's questions through the robot. After the interactive activities, semi-structured interviews were conducted with the children to obtain feedback on their experiences with the robot.

The research team was present during the interaction. In addition to the researchers, who took care of programming the robot, recording the event and logistical issues, a facilitator was also present, who moderated the children's play with the robots, as well as co-led the whole event with NAO.

\section{Data collection and analysis}
We used three cameras and a portable microphone to record the interaction between the children and the robot. Qualitative analysis was conducted using theme categorization and the critical incident technique. Recurring themes in children's behavior, such as body language, facial expressions, and verbal expressions, were coded using MAXQDA software. Robot errors and malfunctions were also coded. The critical incident technique was used to highlight instances of unexpected robot behavior that elicited significant reactions from the children.

\section{Preliminary results}
Based on a qualitative analysis of the video footage and interviews from the events, we classified unexpected robot behaviors into six categories: expressions of the robot's personality, limitations of its movement, inconsistent behavior, mispronunciation, delays, and freezing. The observations from the four sessions were sorted according to three age groups: 6-7 years old, 8-10 years old, and 10-12 years old.

%Though there were four participant groups, data from the 10-12 and 11-12 age groups was combined due to their similarities, resulting in three age groups.
%Though there were four participant groups at the event - a logistical improvement in carrying out interactions in the presence of multiple respondents - three age groups were included in the analysis, combining data from the 10-12 and 11-12 age groups due to the similarities present.
A list of all the robot's unexpected behaviors observed during the study is provided in the appendix.

Children demonstrated varying responses to the robot's errors. As noted by Trunkle et al. \cite{turkle_relational_2006}, \say{Children show perseverance in their efforts to communicate with the robots, including finding ways to explain and excuse the robots' failures to communicate with them.} Our study provides a description of the different styles of perseverance observed in the observed age groups.

\subsection{Children 6-7 yrs}
\subsubsection{Robot's personality}
During the interaction, children were presented with some aspects of the robot's personal profile, including its age, food preferences or knowledge. These expressions were mainly answers to children's questions toward the robot and were mostly spontaneously created by the robot's Wizard-of-Oz operators. 

 The children were also presented with some made-up aspects of the robot's personality, including its family and friends, food preferences, and knowledge.  

Children from this 6-7-year age group were interested in the robot's family and close friends:
\begin{verbatim}
    Child 1: Do you have friends?
    Child 2: Who is your favorite 
    friend?
    Child 3: So do you have a friend?
    NAO: I do.
\end{verbatim}
The children reacted with smiles to the information that NAO has many friends. Later, during the crafting activity, one child created a book for NAO with `mom and dad' drawing so NAO could `recall his own parents'.

The children also engaged in conversations about food with the robot: 
\begin{verbatim}
    Child: What do you eat for supper?
    NAO: Disks.
    Child: And what do you eat for second
    breakfast?
    NAO: Batteries.
\end{verbatim}
To the information about batteries as robot's favorite breakfast, they reacted with smiles and repeating the robot's words to each other.

The children reacted to robot's lack of knowledge about humans.  
\begin{verbatim}
    NAO: What do people eat?
    Child: NAO doesn't know everything! 
    Because he doesn't know what people 
    eat!
\end{verbatim}
The robot's statement led to surprise on children's faces.

\subsubsection{Movement limitations and malfunctions}
Young children also observed robot malfunctions concerning walking or moving its body parts. They explained the lack of movements with the robot's intentionality or cognitive states like `hurting'.
\begin{verbatim}
    Child 1: NAO, do a step forward!
    Child 2: He won't make it because his 
    legs hurt.
    Child 3: Or he doesn't want to!
\end{verbatim}
 
\subsubsection{Inconsistent behaviors}
Children from this 6-7-year age group reacted to inconsistencies in the robot's answers. Once the robot pronounced word `pictures' in a similar way as `kittens': 
\begin{verbatim}
    Facilitator: Come on, let's take 
    some pictures with NAO!
    NAO: Pictures! Pictures!
    Child: <Kittens, kittens>?
\end{verbatim}
The children reacted with surprised facial expressions to this event. 

\subsubsection{Mispronunciation}
Children aged 6-7 yrs reacted to the robot's mispronunciation of words with confused facial expressions, or by directing further questions toward the robot or the human facilitator. One such example is presented above under `inconsistent behavior'. In contrast to the older children, younger children did not make taunting comments about the robot's abilities. 

\subsubsection{Delays}
The children aged 6-7 years did not comment on the delay in robot's answers. They usually reacted by looking away from the robot, starting to wiggle in their seats, or moving away from the robot. 

\subsubsection{Freezing}
Some delays and freezes were expected, but some of them appeared as real malfunctions, where the researchers needed to temporarily distract children or justify the robot's behavior.
Children aged 6-7 years reacted to the robot's freezing by looking away from the robot, starting to wiggle in their seats, or starting to run around.
During one of the freezes, they asked about taking a break for snack:
\begin{verbatim}
    Child: Can we eat something now?
\end{verbatim}

\subsubsection{Perseverance style: nurturing}
The perseverance style of nurturing was characterized as a caring and protective attitude towards the robot. The children treated the robot as if it were a baby, expressing a desire to help it learn and grow. They explained its faulty behavior in a gentle and understanding way, suggesting that the robot was simply tired or needed more time to learn. Both girls and boys exhibited this style of behavior.

\subsection{Children 8-10 yrs}
\subsubsection{Robot's personality}
Children aged 8-10 years were mostly interested in the robot's acquired skills. They challenged the robot's knowledge of mathematics by asking it math-related questions. 
\begin{verbatim}
    Child 1: Do you know math?
    Child 2: I'm sure he doesn't.
    NAO: Yes I do.
    Child 1: So what is 2 + 2?
    NAO: 4
    Children: Wow! 

    Child 1: NAO, what is 5 X 5?
    NAO: 25
    Child: Great!

    Child 1: NAO, count to 10! 
    NAO: *counts from 1 to 10*
    Children: Bravo! 
\end{verbatim}
Children responded to the robot showing its math-related abilities with amazed facial expressions, laughter, clapping hands, or looking surprised at each other. Some of the children video-recorded robot's demonstrations with their phones. 

Children aged 8-10 years were also interested in the robot's language abilities. They asked it if it could speak Spanish, German, Russian and Japanese. 
\begin{verbatim}
    Child: Can you speak Spanish?
    NAO: Si!
    Children: *laugh*
    
    Child 1: Can you speak Japanese?
    NAO: Konnichiwa.
    Child 1: What?!
    Child 2: NAO, you're amazing!
\end{verbatim}
The children reacted to the robot showing off its language abilities with amazed facial expressions, laughter, looking surprised at each other or jumping in their seats. Some of them asked the robot to repeat some sentences in different languages so they could video-record it with their phones. When robot could not say `good morning' in Russian, they reacted with sighs and disappointed facial expressions. 

\subsubsection{Movement limitations and malfunctions}
Children aged 8-10 years challenged the robot's movement abilities as well. 
\begin{verbatim}
    Child: Could you dance hip-hop?
    NAO: I could try.
    Child: Dance, please! 

    Child: Can you walk?
    NAO: Yes. 
    Child: Show us!
    NAO: *makes one little step forward* 
    Children: *look disappointed at each 
    other* 
    Facilitator: Nao has problems with 
    moving sometimes. 

    Child 1: Lay down!
    NAO: *lays down* 
    Child 1: Now stand up! 
    NAO: *stands up* 
    Child 1: So slow...
    Child 2: In this time I would stand 
    up 30 times.
    Facilitator: I think NAO deserves 
    bravos.
    Children: Bravo! 
\end{verbatim}
The children asked NAO many times to dance, walk or jump. They also wanted to interact with it physically, by shaking its hand or doing a high five. They expressed disappointment through their facial expressions when it could not fulfill their requests. Later, during the interviews, they expressed their surprise and disappointment that NAO could not do tricks, jump, or walk that well. However, during the interaction they clapped their hands and video-recorded the robot when it showed off its movements. 

\subsubsection{Inconsistent behaviors}
Some of the robot's behaviors were inconsistent to children aged 8-10 years. 
\begin{verbatim}
    NAO: *plays a sound of ovations and 
    makes a winning pose when a character 
    in the read story wins* 
    Child: Why did he do that? 
    Facilitator: Because this character 
    is happy he won. 
    Child: Oh!
    Children: *clap their hands* 

    NAO: *plays a sound of angry noises 
    when a character in the read story 
    is angry* 
    Child 1: What's going on with him?
    Child 2: I think he's angry.
    Child 3: I think he's drinking 
    something.
\end{verbatim}
The children reacted to the inconsistent behaviors with surprised facial expressions. They started to ask the human facilitator more questions about the situation or discuss it among themselves. They reacted to the robot's `weird' behavior with laughter and clapping hands. Some of them video-recorded it with their phones. 

\subsubsection{Mispronunciation}
There were times during the interaction when the robot mispronounced some words. 
\begin{verbatim}
    Child 1: Do you like cakes?
    NAO: I do.
    Child 2: Oh, he doesn't?
    Child 1: Oh..
    Child 2: Wait, did he say 
    <I don't> or <I do>?

    NAO: *reads a story* "...a machine 
    that could produce things started 
    with the letter 'n'..."
    Child 1: *interrupting NAO* 
    Starting with 'm'?
    Child 2: With 'n'!
    Child 1: Oh.
\end{verbatim}
The robot's mispronunciation led to confusion and surprised facial reactions of children, which slowed the conversation and the story reading. However, it also encouraged them to discuss the situation with each other and ask further questions.

\subsubsection{Delays}
During the interaction, once the robot read a story haltingly: reading few words at a time and stopping for few seconds. Most children still looked at it with interested facial expressions, but some of them started to look away after some time.

\subsubsection{Freezing}
During the interaction, the robot froze few times and was not responding to children's questions. The children patiently asked the robot the same question multiple times, over and over, sometimes by raising their voice. 
\begin{verbatim}
    Child: NAO, sing happy birthday.
    Child: Sing happy birthday.
    Child: Sing happy birthday, 
    please.
\end{verbatim}

\subsubsection{Perseverance style: instrumentalization}
%The perseverance style of instrumentalization was characterized by a determined effort to use the robot as a tool to achieve their goals, despite its limitations.
The children treated the robot as a familiar technology, similar to the voice assistants they use in their daily lives. They interacted with the robot one at a time, taking turns to speak and patiently waiting for a response. They were persistent in their efforts to communicate with the robot, repeating their questions and using simplified language to ensure that it understood them. This behavior reflects a pragmatic and goal-oriented approach to using technology, where the focus is on getting things done efficiently and effectively. The children demonstrated a willingness to work with the limitations of the robot, rather than giving up or becoming frustrated when it did not respond as expected. Overall, the perseverance style of instrumentalization is characterized by a patient and persistent effort to use technology to achieve desired outcomes, despite its limitations.

\subsection{Children 10-12 yrs}
\subsubsection{Robot's personality}
Children aged 10-12 years were interested in the robot's age: 
\begin{verbatim}
    Child: How old are you?
    NAO: I'm six years old.
\end{verbatim}
They reacted by laughing and commenting that the robot is `so young!'. The robot's age was repeatedly mentioned later during the interviews. The children were relating the robot's age to its appearance. One of them suggested that ``NAO is very short for his age''.

The children from this group also asked the robot food-related questions. They started a conversation about hot dogs: 
\begin{verbatim}
    Child: Do you like hot dogs?
    NAO: Yes, but vegan.
    Child: Are you vegan?
    NAO: Yes.
    Child: Why are you vegan?
    NAO: I love animals.
    Child: I love meat.
\end{verbatim}
The children reacted with surprised facial expressions and continued to ask more questions on this topic. They mentioned this later during the interviews by asking the experimenter ``Why is he vegan?''

The children in 10-12 years age group often questioned the robot's knowledge. For example, one child commented:
\begin{verbatim}
    Child: In my opinion robots are 
    stupid until you tell them what 
    to do.
\end{verbatim}
Another child sarcastically replied to the robot's words saying, ``Oh NAO, you're so smart'' and started laughing. 

\subsubsection{Movement limitations and malfunctions}
During the interactions, the robot faced malfunctions related to walking or moving its body parts. 

At the beginning of the interaction, the robot did not move when being invited to come in front of the children. The children started looking at each other and repeatedly said ``He's just shy.'' 
Several times during the interaction, the children asked the robot to perform certain movements: 
\begin{verbatim}
    Child 1: NAO can you dance?
    Child 2: Can you walk?
    Child 1: NAO can you jump? NAO, 
    jump!
    Child 2: Do a back flip!
\end{verbatim}
The children reacted with smiles and applause to the robot sitting (from a standing posture) or dancing. They continued to ask questions about movements even after the robot stated that it could not execute some of them. However, the robot's evasive answers to questions challenging its movement abilities led to taunting comments from these older children. Later in the interviews, the children mentioned disappointment related to the robot's movement abilities and that they would like it to do more `tricks'.

\subsubsection{Inconsistent behaviors}
 The children from the 10-12 years age group reacted with surprise and confusion to inconsistencies in the robot's answers: 
\begin{verbatim}
    Child: He told us he can speak 
    other languages but actually 
    he can't. I asked him to say 'hello'
    in German. At first, he said he knew 
    German. Then he said he didn't know 
    German.
\end{verbatim}
The participants from this group also reacted with surprise to inconsistent behaviors of the robot, such as correctly guessing a child's posture during the game of charades while looking away from the child. The children reacted with surprised facial expressions, hand gestures, and by pointing out this incident to each other.

\subsubsection{Mispronunciation}
The robot mispronouncing a word often during the interaction led to confused facial expressions or further questions to the robot or the human facilitator. 
\begin{verbatim}
    Child 1: What?
    Child 2: Can you repeat?
\end{verbatim}
The children exchanged taunting comments about the robot's abilities referring to such situations. 

\subsubsection{Delays}
During the interactions, sometimes there were delays in the robot answering a question. (These were caused by the slow execution of the robot's software and slow typing speed of the operator.) Often, the children were asking many questions at the same time, which resulted in the robot answering the previous question after a new one was asked. 

The oldest children reacted with surprise, taunting comments, or suspicious looks. 
\begin{verbatim}
    Child: I asked him if he could take 
    over the world and he kind of just 
    nodded. So, I was kind of just 
    terrified, that's why I made 
    a boogie-bomb.
    ...
    Child: Then he said he's vegan. 
    But then I asked him 
    <do you eat meat?> and he said <Yea>. 
    This is sus (suspicious).
\end{verbatim}

\subsubsection{Freezing}
During the interactions, the robot's software crashed a few times resulting in the robot freezing its movement and expression.

During the first workshop with the older children, the robot froze at the beginning of the event when it was still behind the curtain. One of the children shouted to the robot: 
\begin{verbatim}
    Child: Come out, Satan!
\end{verbatim}
Later the child explained its behavior during the interview. 
\begin{verbatim}
    Child: He was behind the curtain, 
    we saw just his eyes, you know, 
    the blue circles. It was staring 
    at us for five minutes. We found it 
    a bit creepy.
\end{verbatim}
\subsubsection{Perseverance style: nurturing}
The perseverance style of nurturing was also observed among children aged 10-12 years. In contrast to the younger groups, only girls exhibited this style of behavior among this group of participants.
\begin{verbatim}
    Child: I have to teach him how 
    to run.
    Experimenter: It wasn't cool that 
    he couldn't walk?
    Child: He can walk, he just doesn't 
    want to.
\end{verbatim}
The girls sat close to the robot, touching it gently and speaking to it softly, as if to allow it a better understanding of the situation. This behavior reflects a nurturing and caring approach to using technology, where the focus is on building a relationship and connection with the robot, rather than simply using it as a tool. The nurturing style was mainly expressed by the female participants.

\subsubsection{Perseverance style: instrumentalization}
Children 10-12 years old exhibited instrumental behavior by persistently addressing the robot to achieve a specific outcome. Despite the robot's limitations, they made several attempts to instruct the robot to perform tasks according to their imagination. This behavior was observed only in a small group of children in this group. The remaining children responded to the robot's limitations with skepticism.

\subsubsection{Perseverance style: skepticism}
The perseverance style of skepticism was characterized by a persistent and questioning attitude towards the robot. When the robot made mistakes, the children grew skeptical of its abilities and reliability. They responded by making jokes about the robot, and by poking fun at its errors and limitations. 

However, despite their skepticism, the children did not stop interacting with the robot. Instead, they continued to talk to it while maintaining a critical and questioning attitude towards its capabilities. 
The children demonstrated a willingness to challenge the robot's authority and question its abilities, while still interacting with it.

\section{Discussion and Conclusions}
\subsection{Unexpected behavior on child-robot interaction}
Our study revealed that the children's reactions to delays varied depending on their age. Children aged 6-7 years old tended to lose interest quickly and shift their gaze away from the robot after waiting for a short while. On the other hand, older children were more persistent and patient in waiting for the robot's response, and they would repeatedly ask it questions. The oldest children, aged 10-12 years, were sometimes skeptical and rated the robot's speed poorly. Inconsistent behaviors from the robot, prolonged gaze, and freezing, generated fear and feelings of unsafe interaction in older children (10-12 yrs), but not in younger children (6-10 yrs). Our study supports previous research suggesting that older children may fear robots due to their appearance, unpredictability, and lack of control or transparency \cite{Rubegni2022, yip_laughing_2019}. We also found that Polish children, who have not been previously studied, also experience fear and an unsafe environment when interacting with robots. In terms of the robot's personality, we found that it resonated well with all age groups. The youngest children were most interested in relationships and preferences, which may be interpreted as children of this age being close with their family and the importance of their desires \cite{CA_development}. Children aged 8-10 were interested in acquired skills like math or singing as they are at the age of becoming more competitive and interested in acquiring new skills. Finally, 10-12 year old children were interested in self-related information like gender and age, as well as the robot's preferences. The interest in the robot's gender might be due to the children's development of their self-knowledge as well as an interest in gender-related issues \cite{eccles_development_1999}\cite{CA_development}.

\subsection{Perseverance styles across different age groups}
We observed varied perseverance styles among three age groups of children, influenced by their cognitive and cultural development. Children aged 6-7 years are starting to develop their abilities of reasoning and abstract thinking, and base their behavior largely on what they have learned from their parent-child relations \cite{eccles_development_1999}. This might be the reason why they exhibited nurturing, treating the robot like a baby, and shyness in some cases as their reaction to robot errors. Older children, aged 8-10, showed instrumentalization, indicating that they are developing problem-solving skills and using the robot as a means to achieve desired outcomes \cite{CA_development}. The oldest age group (10-12 yrs) displayed skepticism, nurturing, and instrumentalization, reflecting their developed reasoning and abstract thinking, ability to understand others' behavior, and capacity to plan and coordinate actions. Their developed cognitive abilities may have led to both questioning the robot's abilities and showing a protective attitude towards it \cite{eccles_development_1999}.  

\subsection{Designing age-appropriate child-robot interactions} 
%%The robot's unexpected behaviors like mispronunciation and inability to perform movements may increase interest in interactions with 6-7-year-old children, but too many malfunctions could lead to boredom and decreased participation. Creating a personal profile of the robot may increase their interest. For 8-10-year-olds, rare unexpected behaviors are okay, but frequent ones may lead to boredom. They are interested in the robot's skills, so incorporating them in the interaction or using the robot as a teacher is suggested. Designers should create physically interactive robots that can perform different movements to attract children. In a school environment, errors that cause fear or undermine the robot's knowledge may hinder interactions with 10-12-year-olds. Designers should avoid behaviors that provoke suspicion or fear, like prolonged gazing and distrustful comments, and create robots with developed personalities to diversify school activities, while avoiding errors that may provoke skepticism.

\subsubsection{Designing interactions appropriate for 6-7-year-old children}
Some of the robot's unexpected behaviors usually considered as errors, like the inability to perform certain movements, inconsistency in replying, and mispronunciation of words, can increase interest in the interaction of children of 6-7 yrs and do not have to be avoided by designers, if they appear rarely. 

However, if the robot's malfunctions occur frequently, such as delays in response or freezing, this could lead to boredom and distractions, with the children eventually losing interest in the interaction. Such events could lower the participation rate in class activities, and this should be taken into consideration. 

Another suggestion for the designers is to create a personal profile of the robot, which may increase the children's interest in participating in class activities involving the robot. Younger children seem to be especially interested in the robot's personal life, relationships and preferences such as favourite meals. 

\subsubsection{Designing interactions appropriate for 8-10-year-old children}
For children between 8 and 10 years old, unexpected robot behaviors such as mispronunciation of words, delays and freezing can become an issue and lead to confusion or boredom when occurring often. However, when they happen rarely they do not seem to be a threat to the interaction because of children's patient way of treating the robot. 

Children this age also seem to be especially interested in the robot's acquired skills, e.g. its mathematical or linguistic abilities. A suggestion to designers would be to incorporate such traits in the interaction, or to exploit robots as math and languages teachers.

Another characteristic the children displayed was their expectation that the robot is mobile and physically interactive. A proposition for designers would be to create robots that can perform different movement abilities, such as dancing or showing different poses, that attract the children. 

\subsubsection{Designing interactions appropriate for 10-12-year-old children}
Our study found that in a school environment, errors that undermine the robot's knowledge or cause fear may hinder smooth child-robot interactions, which could affect the children's participation in social experiences and knowledge acquisition.

To design effective child-robot interactions, designers should take care to avoid behaviors that evoke suspicion or fear. We identified two factors that may cause uneasy feelings in children: prolonged gazing and distrustful comments. To avoid prolonged gazing, designers should follow a gaze aversion pattern similar to that of human-human conversations \cite{acarturk_gaze_2021}. To avoid distrustful comments, designers may develop a quick response system to avoid delays in answering and eventual misunderstandings.

We also found that older children react positively to robots with developed personalities, which could be used to diversify school activities with robots. However, designers should avoid errors like mispronunciation of words, constant movement malfunctions, or inconsistent answers, which may provoke skepticism towards the robot's abilities or intelligence in older children.
\subsection{Limitations and future research}
This study was conducted in Poland with mostly Polish children attending a private international school or extracurricular activities in a museum. These groups of children come from a richer socioeconomic background. Future research needs to explore the validity of design implications in other socioeconomic and cultural contexts, as people from different cultures react differently to robots\cite{evers_relational_2008}. 

%\addtolength{\textheight}{-20cm}   % This command serves to balance the column lengths

%%%%%%%%%%%%%%%%%%%%%%%%%%%%%%%%%%%%%%%%%%%%%%%%%%%%

\section*{ACKNOWLEDGMENT}
This research was supported in part by a grant from the Priority Research Area DigiWorld PSP: U1U/P06/NO/02.19 under the Strategic Programme Excellence Initiative at the Jagiellonian University, and by the National Science Centre, Poland, under the OPUS call in the Weave programme under the project number K/NCN/000142.

\bibliographystyle{IEEEtran}
\bibliography{export}

\appendices
\section*{APPENDIX}

\bgroup
\def\arraystretch{1.6}
\begin{table}[H]
 \centering
  \caption{Unexpected behaviors: children (6-7 yrs)}
  \label{tab1}

\begin{tabularx}{\columnwidth}{X}

\hline
\rowcolor{lightgray}
Category of behavior: Robot's personality\\
\hline
Robot presents information about its close relations, by saying that it has friends or that it doesn't remember its parents.\\
%%\cline{2-2}
Robot presents information about its food preferences by saying that it eats disks for supper and batteries for breakfast. \\
%%\cline{2-2}
Robot presents information about its lack of knowledge by asking children questions.\\
%%\cline{2-2}
Robot presents information about its origin by saying that it's from Japan. \\

\hline
\rowcolor{lightgray}
Category of behavior: Movement limitations\\
\hline
Robot can't walk multiple times during the interaction. \\
%%\cline{2-2}
Robot can't move its hands when asked to. \\
%%\cline{2-2}
Robot can't turn around and asks a facilitator to turn it around multiple times during the interaction.\\
\hline
\rowcolor{lightgray}
Category of behavior: Inconsistent behaviors\\
\hline
Robot reacts to the facilitator's invitation to take pictures by saying `pictures', which to children sounds like `kittens'. \\

\end{tabularx}
\end{table}
\egroup

\bgroup
\def\arraystretch{1.6}
\begin{table}[H]
 \centering
\begin{tabularx}{\columnwidth}{X}

\hline
\rowcolor{lightgray}
Category of behavior: Mispronunciation\\
\hline
Robot reacts to the facilitator's invitation to take pictures by saying `pictures', which to children sounds like `kittens'.\\
\hline
\rowcolor{lightgray}
Category of behavior: Delays\\
\hline
Robot provides answers after relatively long time from the questions being asked (relating to human answering questions). \\

\hline
\rowcolor{lightgray}
Category of behavior: Freezing\\
\hline
Robot freezes multiple times during the interaction and doesn't move or speak.\\
%%\cline{2-2}
Robot freezes multiple times during the interaction and doesn't respond to children's or facilitator's questions.\\
\end{tabularx}
\end{table}
\egroup

\bgroup
\def\arraystretch{1.6}
\begin{table}[H]
 \centering
  \caption{Unexpected behaviors: children (8-10 yrs)}
  \label{tab2}

\begin{tabularx}{\columnwidth}{X}

\hline
\rowcolor{lightgray}Category of behavior: Robot's personality\\
\hline
Robot presents information about its math abilities, correctly solving a math problem `what is 2+2'.\\
%%\cline{2-2}
Robot presents information about its math abilities, correctly solving a math problem `what is 5x5'.\\
%%\cline{2-2}
Robot presents information about its math abilities, by correctly counting from 1 to 10.\\
%%\cline{2-2}
Robot presents information about its language abilities by saying `si' in Spanish.\\
%%\cline{2-2}
Robot presents information about its language abilities by saying `konnichi'wa' in Japanese.\\
%%\cline{2-2}
Robot presents information about its language abilities by saying `spasiba' in Russian.\\

\hline
\rowcolor{lightgray}Category of behavior: Movement limitations\\
\hline
 Robot can't walk more than few steps. \\
%%\cline{2-2}
Robot is slow in moving and changing poses, e.g. sitting or standing up. \\
%%\cline{2-2}
Robot can't move its hand to shake hands with children. \\
%%\cline{2-2}
Robot can't move its hand to do a high-five with children. \\
%%\cline{2-2}
Robot can't jump. \\
%%\cline{2-2}
Robot can't dance certain dances. \\
\hline
\rowcolor{lightgray}Category of behavior: Inconsistent behaviors\\
\hline
Robot plays growling sound while playing a character in a story. \\
%%\cline{2-2}
Robot plays ovation sound while playing a character in a story. \\
%%\cline{2-2}
Robot plays machine sound while playing a character in a story. \\
%%\cline{2-2}
Robot plays automatic information about low battery while playing a character in a story.\\
\hline
\rowcolor{lightgray}Category of behavior: Mispronunciation\\
\hline
Robot says ``I do'' that sounds like ``I don't''.\\
%%\cline{2-2}
Robot says `n' that sounds like `m'.\\

\hline
\rowcolor{lightgray}Category of behavior: Delays\\
\hline
Robot reads a story by reading few words at a time and stopping for few seconds. \\
\hline
\rowcolor{lightgray}Category of behavior: Freezing\\
\hline
Robot freezes multiple times during the interaction and doesn't respond to children's questions and requirements. 
\end{tabularx}
\end{table}
\egroup

\bgroup
\def\arraystretch{1.6}
\begin{table}[H]
 \centering
  \caption{Unexpected behaviors: children (10-12 yrs)}
  \label{tab3}

\begin{tabularx}{\columnwidth}{X}

\hline
\rowcolor{lightgray}Category of behavior: Robot's personality\\

\hline
Robot presents information about its age, by saying ``I'm six years old.''\\
%\cline{2-2}
 Robot presents information about its food preferences, by saying that it likes vegan hotdogs and that it's vegan. \\
%\cline{2-2}
 Robot presents its knowledge by answering children's questions.\\
 %\cline{2-2}
 Robot presents information about its belonging by saying ``I am an owner to myself.''\\
%\cline{2-2}
 Robot presents information about its origin by saying that it's from Japan.\\
 %\cline{2-2}
 Robot presents information about its gender by saying that it's a boy.\\
 \hline
 \rowcolor{lightgray}Category of behavior: Movement limitations\\
\hline
 Robot can't walk while being invited to meet the children at the beginning of the interaction and has to be carried by the facilitator. \\
 %\cline{2-2}
 Robot can't turn around and asks a facilitator to turn it around multiple times during the interaction. \\ 
 %\cline{2-2}
 Robot can't dance a certain dance while being asked to.\\
 %\cline{2-2}
 Robot can't run while being asked to.\\
 %\cline{2-2}
 Robot can't jump while being asked to.\\
 \hline
 \rowcolor{lightgray}Category of behavior: Inconsistent behaviors\\
\hline
 Robot tells children it knows German language and later it tells children it doesn't know German language.\\
 %\cline{2-2}
 Robot is correctly guessing child's pose during the game of charades while having its face pointed in opposite direction than the child.\\
 %\cline{2-2}
 Robot tells children that it's vegan and later on nods while a child asks it if it eats meet.\\
 %\cline{2-2}
 Robot says it will dance and doesn't dance.\\
 \hline
 \rowcolor{lightgray}Category of behavior: Mispronunciation\\
\hline
 Robot says ``I will in a minoote'' instead of `a minute'.\\
 %\cline{2-2}
 Robot reacts to a question ``Why are you vegan?'' by saying ``I love animals'', which to children sounds as ``I love lemons''.\\

 \hline
 \rowcolor{lightgray}Category of behavior: Delays\\
\hline
 Robot provides answers after relatively long time from the questions being asked (relating to human answering questions).\\
 %\cline{2-2}
 Robot answers a previous question by saying `yes' while a child asks a new one: ``Do you eat meat?'' \\
 %\cline{2-2}
 Robot answers a previous question by nodding while a child asks a new one: ``Can you take over the word?''\\
 \hline
 \rowcolor{lightgray}Category of behavior: Freezing\\
\hline
 Robot freezes multiple times during the interaction and doesn't move or speak.\\
 %\cline{2-2}
Robot freezes multiple times during the interaction and doesn't respond to children's or facilitator's questions.\\

\end{tabularx}
\end{table}
\egroup

\end{document}